\documentclass{llncs}
\usepackage{times}
\usepackage{graphicx}
\usepackage{latexsym}
\usepackage{subfigure}
\usepackage{xspace}
\begin{document}

\newtheorem{mydefinition}{Definition}
\newtheorem{mytheorem}{Proposition}
\newtheorem{myexample}{}{\bf}{\it}
\newtheorem{mytheorem1}{Theorem}
\newcommand{\myproof}{\noindent {\bf Proof:\ \ }}
\newcommand{\myqed}{\mbox{$\Box$}}
\newcommand{\myend}{\mbox{$\clubsuit$}}

\newcommand{\doublelex}{\mbox{\sc DoubleLex}}

\newcommand{\dyn}{\mbox{\sc Dyn}}
\newcommand{\preced}{\mbox{\sc Prec}}
\newcommand{\prep}{\mbox{\sc Prep}}
\newcommand{\precedsh}{\mbox{\sc $\preced_{sh}$}}
\newcommand{\precedprep}{\mbox{\sc $\preced+\prep$}}
\newcommand{\precedprepsh}{\mbox{\sc $\preced+\prep_{sh}$}}

\title{On the Complexity of Breaking Symmetry}

\author{Toby Walsh}
\institute{NICTA and UNSW, Sydney, Australia,
email: Toby.Walsh@nicta.com.au}


\maketitle
\bibliographystyle{splncs}

\begin{abstract}
We can break symmetry
by eliminating solutions within a symmetry class
that are not least in the lexicographical ordering. 
This is often referred to as the lex-leader method. 
Unfortunately, as symmetry groups can be large,
the lex-leader method is not tractable in general. 
We prove that using other total orderings besides
the usual lexicographical ordering will not reduce the computational
complexity of breaking symmetry in general. 
It follows that breaking symmetry with other orderings
like the Gray code ordering or the Snake-Lex
ordering is intractable in general. 
\end{abstract}

\section{Introduction}

Symmetry occurs in many combinatorial problems
(e.g. if trucks in a routing problem are located
at the same depot and have the same capacity, they
may be interchangeable in any solution). 
Symmetry can also be introduced by modelling
decisions (e.g. using a set of finite domain variables
to model a set of objects can introduce symmetries that
permute these variables). 
A common method to deal with symmetry is to
add constraints which eliminate symmetric
solutions
\cite{puget:Sym,ssat2001,ffhkmpwcp2002,llconstraints06,wecai2006,wcp06,wcp07,llwycp07,waaai2008,knwcp10}.
Crawford {\it et al.} have proved
that breaking symmetry by adding constraints
to eliminate symmetric solutions
is intractable in general.
\cite{clgrkr96}.
More specifically, they prove that, for a matrix
model with row and column
symmetries, deciding if an assignment is the 
smallest in its symmetry class is NP-hard where
we append rows together 
and compare them lexicographically.  
There is, however, nothing 
special about appending rows together or
comparing assignments lexicographically. We 
could use any total ordering
over assignments. 

For example, we could break symmetry 
with the Gray code ordering. That is,
we add constraints that eliminate symmetric
solutions within each symmetry 
class that are not smallest in the Gray code ordering. 
The Gray code ordering is a total ordering over
assignments used in error correcting codes. For instance, the
4-bit Gray code orders assignments as follows:
\begin{eqnarray*}
&0000,
0001,
0011,
0010,
0110,
0111,
0101,
0100,& \\
&\ \ \ \ \ \ \ \ 1100,
1101,
1111,
1110,
1010,
1011,
1001,
1000
&
\end{eqnarray*}
Such an ordering will pick out different
solutions in each symmetry class. 
We consider
here the binary Gray code but note that it can be generalized
to deal with non-binary domains. 
The Gray code ordering has some properties that
may make it useful for symmetry breaking. In particular,
neighbouring assignments in the ordering only differ
at one position, and flipping just one bit reverses
the ordering of the subsequent bits. 
As a second example, we could break row and
column symmetry
in a matrix model with the Snake-Lex ordering \cite{snakelex}.
This orders assignments within a symmetry
class by lexicographically comparing 
vectors constructed by appending the entries in the matrix
in a ``snake like'' manner in which the
first column is appended to the reverse
of the second column, and this is then
appended to the third column, and then the
reverse of the fourth column and so on. 
Again, this picks out different solutions
in each symmetry class.

Can we be sure that different
orderings like the Gray code and Snake-Lex
ordering do not change the computational complexity
of breaking symmetry? 
In this paper, we argue that breaking 
symmetry with a different ordering over
assignments is unlikely to improve the complexity.
Our argument breaks into two parts. First, we argue
that using a different ordering can increase the computational
complexity of breaking symmetry. 
Second, we argue that, under modest assumptions (which
are satisfied by the Gray code and Snake-Lex orderings),
we cannot reduce the computational complexity from that
of breaking symmetry with a lexicographical ordering.
Many dynamic methods for dealing with symmetry
are equivalent to posting symmetry breaking
constraints ``on the fly'' (e.g. 
\cite{bscade92,bsjar94,backofen:Sym,sbds,dynamiclex,kwecai08,kwecai10}. 
Hence, our results are likely to have implications for such
dynamic methods too. 

\section{Background}

A symmetry of a set of constraints $S$ is a bijection $\sigma$
on variable-value pairs 
that maps solutions onto solutions \cite{cjjpsconstraints06}.
We lift $\sigma$ from variable-value pairs
to complete assignments (and hence solutions) in the 
natural way: $\sigma(A) = \{ \sigma(X=v) \ | \ X=v \in A\}$. 
Thus, a bijection $\sigma$ on variable-value pairs
is a {\em symmetry} of a set of constraints 
$S$ iff given any solution $A$ of $S$, $\sigma(A)$ is also a solution
of $S$. 
A {\em variable symmetry} is a bijection that just
acts on the variable indices, whilst a {\em value symmetry}
is a bijection that just acts on the values.
The set of symmetries form a group under composition.
Given a symmetry group $\Sigma$, a subset $\Pi$
generates $\Sigma$ iff any $\sigma \in \Sigma$ is a composition
of elements from $\Pi$. 
A symmetry group $\Sigma$ partitions the solutions
into symmetry classes (or orbits). We write $[A]_{\Sigma}$ for
the symmetry class of solutions symmetric to
the solution $A$. Where $\Sigma$ is clear from the 
context, we simply write $[A]$. Note that symmetry classes
are equivalence classes. 
A set of symmetry
breaking constraints is {\em sound} iff it
leaves at least one solution in each symmetry
class, and {\em complete} 
iff it leaves at most one solution in each symmetry
class. 

We will study what happens to symmetries when problems
are reformulated onto equivalent problems. For example, we 
might consider the Boolean form of a problem in which we map
$X_i=j$ onto $Z_{ij}=1$. 
Two sets of constraints, $S$ and $T$ over
possibly different variables
are {\em equivalent} iff there is a bijection $\pi$ between
solutions of $S$ and of $T$. 
Suppose $U_i$ and $V_i$ for $i \in [1,k]$ are partitions of
the sets $U$ and $V$ into $k$ subsets. 
Then the two partitions are {\em isomorphic}
iff there are bijections $\pi: U \mapsto V$ 
and $\tau:[1,k]\mapsto [1,k]$ such
that $\pi(U_i)=V_{\tau(i)}$ for $i \in [1,k]$
where $\pi(U_i) = \{ \pi(u) \ | \ u \in U_i \}$.. 
Two groups of symmetries $\Sigma$ and $\Pi$
of constraints $S$ and $T$ respectively are {\em isomorphic} iff
$S$ and $T$ are equivalent, and
their symmetry classes of solutions are isomorphic.
When two groups of symmetries are isomorphic, 
the number and sizes of their symmetry
classes are identical. 

\section{Using other orderings}

Crawford {\it et al.} proposed the lex-leader method, 
a general way to break symmetry statically using lexicographical
ordering constraints \cite{clgrkr96}.
This method picks
out the lexicographically smallest solution in each
symmetry class. To do this, 
we post a lexicographical ordering constraint
for every symmetry $\sigma$:
$$ \langle X_1, \ldots, X_n \rangle \leq_{\rm lex} 
\sigma (\langle X_1, \ldots, X_n \rangle) 
$$
Where $X_1$ to $X_n$ is some ordering on the 
variables in the problem. 
Many static symmetry breaking constraints 
can be derived from such lex-leader constraints.
For example, {\sc DoubleLex} constraints
to break row and column symmetry can be 
derived from them \cite{lex2001}. 
Efficient algorithms have
been developed to propagate
such static symmetry
breaking constraints
(e.g. \cite{fhkmwcp2002,fhkmwaij06,knwercim09}).

We first argue that other
orderings besides the lexicographical ordering
can increase the computational
complexity of symmetry breaking
even when we have just a single symmetry
to eliminate. Our argument has two
parts. We first argue that, even with 
a single symmetry, we can introduce
computational complexity through
the complexity of deciding the ordering. We then argue that,
when the ordering is polynomial to decide,
we can introduce computational complexity 
by having many symmetries to break. 

\begin{mytheorem}
There exists a total ordering $\preceq$ on assignments,
and a class of problems $P$ that is polynomial to decide
and has a single symmetry $\sigma$ such
that finding a solution of $\{ S \cup 
\{ \langle X_1, \ldots, X_n \rangle \preceq
\sigma (\langle X_1, \ldots, X_n \rangle) \} \ | \ S \in P\}$
is NP-hard,
whilst finding a solution of $\{ S \cup 
\{ \langle X_1, \ldots, X_n \rangle \leq_{\rm lex}
\sigma (\langle X_1, \ldots, X_n \rangle) \} \ | \ S \in P\}$
is polynomial. 
\end{mytheorem}
\myproof
Reduction from 1-in-3-SAT on $m$ positive
clauses. In 1-in-3-SAT, we wish to decide
if 3-cnf formula can be satisfied by a truth
assignment that sets exactly 1 out of the 3 literals
in each clause to true. Let 
$n=3m+1$, 
and $S$ be a set of 
unary constraints that ensure $X_{3p-2}=i$, $X_{3p-1}=j$, $X_{3p}=k$ 
where $p \in [1,m]$ and the $p$th clause is $i \vee j \vee k$.
We also have $X_{3m+1} \in \{0,1\}$. 
Consider the symmetry $\sigma$ that
interchanges the values $0$ and $1$ for
$X_{3m+1}$. 
We define an ordering $\prec$ as follows:

$\langle X_1, \ldots, X_{3m+1} \rangle
\prec \langle Y_1, \ldots, Y_{3m+1} \rangle$
iff one of 3 conditions holds:
\begin{enumerate}
\item $\langle X_1, \ldots, X_{3m} \rangle
<_{\rm lex} \langle Y_1, \ldots, Y_{3m} \rangle$,
or 
\item
$\langle X_1, \ldots, X_{3m} \rangle
= \langle Y_1, \ldots, Y_{3m} \rangle$,
$X_{3m+1}=0$, $Y_{3m+1}=1$, 
and the 1 in 3-SAT problem defined
by $X_1$ to $X_{3m}$ is satisfiable,
or 
\item
$\langle X_1, \ldots, X_{3m} \rangle
= \langle Y_1, \ldots, Y_{3m} \rangle$,
$X_{3m+1}=1$, $Y_{3m+1}=0$,
and the 1 in 3-SAT problem defined
by $X_1$ to $X_{3m}$ is unsatisfiable.
\end{enumerate}
Finally, we define $\preceq$ by
$\langle X_1, \ldots, X_{3m+1} \rangle
\preceq \langle Y_1, \ldots, Y_{3m+1} \rangle$
iff 
$\langle X_1, \ldots, X_{3m+1} \rangle
\prec \langle Y_1, \ldots, Y_{3m+1} \rangle$
or
$\langle X_1, \ldots, X_{3m+1} \rangle
= \langle Y_1, \ldots, Y_{3m+1} \rangle$. 
Now, the (only) solution of 
$S \cup 
\{ \langle X_1, \ldots, X_n \rangle \preceq
\langle \sigma(X_1), \ldots, \sigma(X_n) \rangle \}$
has $X_{3m+1}=0$ if the corresponding
1 in 3-SAT problem is satisfiable
and $X_{3m+1}=1$ otherwise. Hence, 
finding the solution of these constraints
is NP-hard. By comparison, 
finding a solution of $S \cup 
\{ \langle X_1, \ldots, X_n \rangle \leq_{\rm lex}
\langle \sigma(X_1), \ldots, \sigma(X_n) \rangle \}$ is
polynomial as it 
always has the solution in which
$X_{3m+1}=0$. 
\myqed

In the example in this proof, 
the ordering used to break symmetry was 
computationally intractable to decide. 
More precisely, deciding if 
$\langle X_1, \ldots, X_{n} \rangle
\preceq \langle Y_1, \ldots, Y_{n} \rangle$ was
NP-hard. 
If we insist that the ordering
used to break symmetry is polynomial to decide,
then breaking a single symmetry will also be polynomial.
Indeed, breaking even a polynomial number of 
symmetries must also be polynomial in this case. 
However, if we have an exponential number of symmetries
to break then changing the ordering used
to break symmetry from the lexicographical
ordering to some other ordering can increase
the computational complexity of finding a solution. 

\begin{mytheorem}
There exists
a total ordering $\preceq$ on assignments where 
deciding $\preceq$ is polynomial, and
a class of problems $P$ that is also polynomial to decide,
and for each $S \in P$, a symmetry group $\Sigma$ of $S$
such that finding a solution of 
$\{ S \cup 
\{ \langle X_1, \ldots, X_n \rangle \preceq
\sigma (\langle X_1, \ldots, X_n \rangle) \ | \ \sigma \in \Sigma \} \ | \ S \in P\}$
is NP-hard,
but finding a solution of $\{ S \cup 
\{ \langle X_1, \ldots, X_n \rangle \leq_{\rm lex}
\sigma (\langle X_1, \ldots, X_n \rangle) \ | \ \sigma \in \Sigma \} \ | \ S \in P\}$
is polynomial. 
\end{mytheorem}
\myproof
Reduction from SAT. 
We let $\preceq$ be the reverse lexicographical
ordering, $\geq_{\rm lex}$. This is clearly
polynomial to decide. 
Consider any SAT formula $\varphi$ on 0/1 variables 
$X_1$ to $X_n$. 
Let $S$ be $\varphi \vee (X_1=\ldots =X_n=0)$,
and the symmetry group $\Sigma$ be such that 
all solutions of $S$ are in the same symmetry class.
Consider any solution of 
$S \cup 
\{ \langle X_1, \ldots, X_n \rangle \preceq
\sigma(\langle X_1, \ldots, X_n \rangle) \ | \ \sigma 
\in \Sigma \}$. 
There are three cases. In the first case,
the solution is $X_1=\ldots =X_n=0$ and this is not 
a solution of $\varphi$. Then $\varphi$ is
unsatisfiable. 
In the second case, 
the solution is $X_1=\ldots =X_n=0$ and this is 
the only solution of $\varphi$. Then $\varphi$ is
satisfiable.
In the third case, 
the solution is lexicographical larger
than $X_1=\ldots =X_n=0$. Then $\varphi$ is
again satisfiable. Hence finding a solution
of $S \cup 
\{ \langle X_1, \ldots, X_n \rangle \preceq
\sigma (\langle X_1, \ldots, X_n \rangle) \ | \ \sigma 
\in \Sigma \}$
decides the satisfiability of $\varphi$. Thus
finding a solution is NP-hard. 
By comparison, 
finding a solution of $S \cup 
\{ \langle X_1, \ldots, X_n \rangle \leq_{\rm lex}
\sigma(\langle X_1, \ldots, X_n \rangle) \ | \ \sigma 
\in \Sigma \}$
is polynomial as it 
always has the solution in which
$X_1=\ldots = X_n=0$. 
\myqed

A criticism that can be made 
of the examples in the last two proofs is that
both are rather ``artificial''. In the first proof, we 
introduced complexity into symmetry breaking
by making the ordering NP-hard to decide. In the 
second proof, we 
introduced complexity into symmetry breaking
by making the symmetry group NP-hard to decide. 
In the rest of this paper, we give a 
more natural example to show that symmetry
breaking with other orderings is intractable. 
Our argument uses a symmetry group whose
elements are easy to generate and which is isomorphic to the 
symmetry group that interchanges rows and columns in
a matrix model. In addition, we suppose that the ordering
is polynomial to decide so cannot itself be a source
of computational complexity. 

\section{Symmetry under reformulation}

We will show that, under some modest assumptions,
we cannot pick an ordering with which to break
symmetry that will make it computationally easier than using
the simple lexicographical ordering.
Our argument breaks into two parts. First,
we show how the symmetry of a problem changes
when we reformulate onto 
an equivalent problem. 
Second, we argue that we can map 
symmetry breaking with any other ordering 
onto symmetry breaking using the lexicographical
ordering on an equivalent problem. 
Therefore, breaking symmetry with a different
ordering cannot have a lesser
computational complexity. 
We begin by proving that 
reformulation maps the symmetry
group of a problem onto an isomorphic
symmetry group. 

\begin{mytheorem}
If a set of constraints $S$ has a symmetry group
$\Sigma$, $S$ and
$T$ are equivalent sets of constraints,
$\pi$ is any bijection between solutions
of $S$ and $T$, and $\Pi \subseteq \Sigma$ then:
\begin{enumerate}
\item[(a)] $\pi \Sigma \pi^{-1}$ is
a symmetry group of $T$;
\item[(b)] $\Sigma$ and $\pi \Sigma \pi^{-1}$ are 
isomorphic symmetry groups;
\item[(c)] if $\Pi$ generates $\Sigma$ then
$\pi \Pi \pi^{-1}$ generates $\pi \Sigma \pi^{-1}$.
\end{enumerate}
\end{mytheorem}
\myproof
(a) Consider any solution $A$ of $T$ and any $\sigma \in \Sigma$. Then
$\pi^{-1}(A)$ is a solution of $S$. As $\sigma$ is a symmetry
of $S$, $\sigma(\pi^{-1}(A))$ is a solution of $S$. Hence, 
$\pi(\sigma(\pi^{-1}(A)))$ is a solution of $T$. Thus,
$\pi \sigma \pi^{-1}$ is a symmetry of $T$. 
Hence $\pi \Sigma \pi^{-1}$ is a symmetry group of $T$. 

(b) $S$ and $T$ are equivalent sets of 
constraints. Consider any solution $A$ of $S$.
Then the bijection $\pi$ maps the symmetry class
$[A]_{\Sigma}$ onto the isomorphic
symmetry class $[\pi(A)]_{\pi \Sigma \pi^{-1}}$.
Consider two symmetric solutions $B$ and $C$ from 
$[A]_{\Sigma}$ where $B \neq C$. 
As they are in the same symmetry class,
there exists $\sigma \in \Sigma$
with $\sigma(B)=C$. 
The bijection $\pi$ maps
$B$ and $C$ onto 
$\pi(B)$ and $\pi(C)$ respectively. 
Consider the symmetry $\pi \sigma \pi^{-1}$ in
$\pi \Sigma \pi^{-1}$. 
Now $\pi (\sigma (\pi^{-1} (\pi(B)))) = 
\pi(\sigma(B)) = \pi(C)$. 
Thus $\pi(B)$ and $\pi(C)$ are
in the same symmetry class.
As $\pi$ is a bijection and $B \neq C$, 
it follows that $\pi(B) \neq \pi(C)$. 
Hence, this symmetry class has the same size as 
$[A]_{\Sigma}$. Thus $\Sigma$ and $\pi \Sigma \pi^{-1}$ are 
isomorphic symmetry groups. 

(c) Consider any $\sigma \in \Sigma$. There exist
$\sigma_1$, \ldots, $\sigma_n \in \Pi$ such that
$\sigma$ is generated from the product $\sigma_1 \cdots \sigma_n$. 
Consider $\pi \sigma_1 \pi^{-1}$, \ldots, $\pi \sigma_n \pi^{-1} \in 
\pi \Pi \pi^{-1}$. Their product is 
$\pi \sigma_1 \pi^{-1}  \pi \sigma_2 \pi^{-1} \cdots
\pi \sigma_n \pi^{-1}$ which simplifies to
$\pi \sigma_1 \cdots \sigma_n \pi^{-1}$ and
thus to $\pi \sigma \pi^{-1}$. Hence $\Pi$ 
generates $\pi \Sigma \pi^{-1}$. 
\myqed

We next show that a sound (complete) set of 
symmetry breaking constraints will be mapped
by the reformulation onto a sound (complete) set of 
symmetry breaking constraints for the reformulated
problem. 

\begin{mytheorem}
If a set of constraints $S$ has a symmetry group
$\Sigma$, 
$B$ is a sound (complete) set of symmetry 
breaking constraints for $S$,
and $S$ and $T$ are equivalent sets of constraints,
then $\pi(B)$ is a sound (complete) set of symmetry 
breaking constraints for $T$ 
where $\pi$ is a bijection between solutions
of $S$ and $T$. 
\end{mytheorem}
\myproof
(Soundness)
Suppose $B$ is a sound set of symmetry 
breaking constraints for $S$. 
Consider any $A \in sol(S \cup B)$.
Now $A \in sol(S)$
and $A \in sol(B)$.
But as $\pi$ is a bijection between solutions
of $S$ and $T$
$\pi(A) \in sol(T)$.
Since $A \in sol(B)$, 
it follows that $\pi(A) \in sol(\pi(B))$ \cite{kwecai10}. 
Thus, $\pi(A) \in sol(T \cup \pi(B))$. 
Hence, there is at least one solution left by 
$\pi(B)$ in every symmetry class of $T$. 
That is, $\pi(B)$ 
is a sound set of symmetry breaking constraints.

(Completeness)
Suppose $B$ is a complete set of symmetry
breaking constraints for $S$. 
Consider any $A \in sol(T \cup \pi(B))$
Now $A \in sol(T)$ and $A \in sol(\pi(B))$. 
But as $\pi$ is a bijection between solutions of $S$
and $T$, $\pi^{-1}$ is a bijection between solutions of
$T$ and $S$. Hence $\pi^{-1}(A) \in sol(S)$. 
Since $A \in sol(\pi(B))$, it follows 
that $\pi^{-1}(A) \in sol(B)$. 
Thus $\pi^{-11}(A) \in sol(S \cup B)$. 
Hence, there is at most one solution left by 
$\pi(B)$ in every symmetry class of $S$. 
That is, $\pi(B)$ 
is a complete set of symmetry breaking constraints.
\myqed

We have shown that reformulating onto an equivalent problem just
maps the symmetries onto isomorphic symmetries and
simply requires the same reformulation
of any symmetry breaking constraints. 
We will use these results to argue
that symmetry breaking with any ordering
besides the lexicographical ordering is
intractable for a symmetry group isomorphic to
the symmetry group that permutes rows and columns
in a matrix model.

\section{Breaking symmetry is intractable}

Suppose we break symmetry using some other
ordering than the usual lexicographical ordering
on assignments. For example, suppose we 
break symmetry by insisting that any solution
is the smallest symmetric solution in each
symmetry class under the Gray code ordering. 
Under modest assumptions, we can lower bound the
complexity of symmetry breaking. 
We consider
orderings which are {\em simple}. In such an ordering 
we can compute the position of any assignment in
the ordering in polynomial time, and given
any position in the ordering we can compute the
assignment at this position. 
For example, for 0/1 variables and a lexicographical
ordering, $\langle 0,\ldots,0,0,0 \rangle$ is
in first position in the lexicographical ordering, 
$\langle 0,\ldots,0,0,1 \rangle$ is in
second position, 
$\langle 0,\ldots,0,1,0 \rangle$ is in
third position, 
$\langle 1,\ldots,1,1 \rangle$ is in
$2^n$th (or last) position. 
Given any assignment,
we can compute its position in the lexicographical
ordering in polynomial time.
Similarly, we can compute the $k$th assignment
in the lexicographical ordering in polynomial time.
As a second example
for 0/1 variables and the Gray code
ordering, $\langle 0,\ldots,0,0,0 \rangle$ is
in first position in the Gray code ordering, 
$\langle 0,\ldots,0,0,1 \rangle$ is in
second position, 
$\langle 0,\ldots,0,1,1 \rangle$ is
in third position, 
$\langle 1,0,\ldots,0,0 \rangle$ is
in 
$2^n$th (or last) position, and
we can compute these positions in polynomial time.
Similarly, we can compute the $k$th assignment
in the Gray code ordering in polynomial time.

We now 
give our main result which
generalizes the result in \cite{clgrkr96}
that computing the lex-leader assignment is
NP-hard. We prove that computing the smallest
symmetry of an assignment according
to {\em any} simple ordering is NP-hard.

\begin{mytheorem}
Given any simple ordering $\preceq$,
there exists a symmetry group
such that deciding if
an assignment is smallest in its symmetry
class according to $\preceq$ is NP-hard.
\end{mytheorem}
\myproof
For a $n$ by $n$ 0/1
matrix with row and column symmetry, deciding if
an assignment is smallest in its symmetry
class according to $\leq_{\rm lex}$ is NP-hard
\cite{clgrkr96}. 
Since $\preceq$ and the lexicographical order are both
simple orderings, there exists a polynomial
function $f$ to map assignments onto their
position in the lexicographical ordering,
and a polynomial function $g$ to map 
position in the $\preceq$ ordering
onto the corresponding assignment. 
Consider the mapping $\pi$ 
defined by $\pi(A)=g(f(A))$ for any 
complete assignment $A$. Now $\pi$ is a permutation that
is polynomial to compute which
maps the total ordering of assignments of 
$\leq_{\rm lex}$ onto
that for $\preceq$. 
Similarly, 
$\pi^{-1}$ is a permutation that
is polynomial to compute which
maps the total ordering of assignments of 
$\preceq$ onto
that for $\leq_{\rm lex}$. 
Let $\Sigma_{rc}$ be the row and column symmetry group.
By Proposition 3,
the problem of finding 
the lexicographical least
element of each symmetry class 
for $\Sigma_{rc}$ is equivalent
to problem of finding 
the least
element of each symmetry class 
for $\pi \Sigma_{rc} \pi^{-1}$. 
Thus, 
for the symmetry group
$\pi \Sigma_{rc} \pi^{-1}$
deciding if
an assignment is smallest in its symmetry
class according to $\preceq$ is NP-hard.
\myqed

If follows that even checking a constraint
which decides if an assignment is the smallest
member of its symmetry class according to $\preceq$
is NP-hard. Note that the Gray code ordering
is simple. Hence, a corollary of 
Proposition 5 is that breaking symmetry with the
Gray code ordering is NP-hard in general. 
It also follows from 
Proposition 5 that breaking symmetry with the
Snake-Lex ordering is NP-hard in general. 

\section{Conclusions}

We have argued that breaking 
symmetry with a different ordering over
assignments than the usual lexicographical
ordering used by the lex-leader method
does not improve the complexity of 
dealing with symmetry.
Our argument had two parts. First, we argued
that using a different ordering can increase the computational
complexity of breaking symmetry. 
Second, we argued that, under modest assumptions,
we cannot reduce the computational complexity from that
of breaking symmetry with a lexicographical ordering.
These assumptions are satisfied by the Gray code and 
Snake-Lex orderings. Hence, it follows that these
methods of breaking symmetry are also intractable
in general. 

\section*{Appendix: Gray code constraint}

To demonstrate that we could break symmetry 
efficiently with a Gray code
ordering,  we give an encoding for the constraint
$Gray([X1,\ldots,Xn],[Y1,\ldots,Yn])$
that ensures $\langle X1,\ldots,Xn \rangle$ is before  $\langle 
Y1,\ldots,Yn \rangle$
in the Gray code ordering where $X_i$ and $Y_j$ are 0/1 finite domain
variables. For a variable symmetry, we need just then
set $Y_i = X_{\sigma(i)}$ where $\sigma$ is an appropriate
bijection on variable indices,
whilst for a value symmetry, we need just set
$Y_i = \theta(X_i)$ where $\theta$ is an appropriate
bijection on values.

We suppose the existence of a sequence of 0/1/-1 state variables, $Q_1$ 
to $Q_{n+1}$.
We encode the Gray code constraint
by means of the following decomposition
where $1 \leq i \leq n$:
\begin{eqnarray*}
&Q_1 =  1& \\
&Q_{n+1} =  0& \\
&Q_i \neq 1  \vee  X_i \leq Y_i& \\
&Q_i \neq -1 \vee  X_i \geq Y_i& \\
&X_i = Y_i \vee  Q_{i+1}=0& \\
&X_i = 1 \vee Y_i = 1 \vee Q_{i+1}=Q_i &\\
&X_i = 0 \vee Y_i = 0 \vee Q_{i+1}=-Q_i &
\end{eqnarray*}
We can show that this decomposition does not hinder propagation.

\begin{mytheorem}
Unit propagation on this decomposition
achieves domain consistency 
on $Gray([X1,\ldots,Xn],[Y1,\ldots,Yn])$
in $O(n)$ time.
\end{mytheorem}
\myproof
(Correctness)
$Q_i$ is set to 0 as soon as the 
two vectors are ordered correctly. 
$Q_i$ is set to 1 iff the $i$th bits,
$X_i$ and $Y_i$ 
are ordered in the Gray code
ordering with 0 before 1. $Q_i$ is 
set to -1 iff the $i$th bits,
$X_i$ and $Y_i$ 
are ordered in the Gray code
ordering with 1 before 0.
$Q_{i+1}$ stays the same polarity
iff $X_i=Y_i=0$ and flips 
polarity iff $X_1=Y_i=1$. 

(Completeness)
The decomposition is Berge acyclic. Thus
unit propagation is enough to guarantee 
the existence of a support for every value. 

(Complexity)
There are $O(n)$ disjuncts in the decomposition.
Hence unit propagation takes $O(n)$ time. 
\myqed

In fact, it is possible to show that 
the {\em total}
time to enforce domain consistency
at each branching decision 
down a branch of the search tree is $O(n)$. 

\bibliography{/Users/twalsh/Documents/biblio/a-z2,/Users/twalsh/Documents/biblio/pub2}

\begin{thebibliography}{10}

\bibitem{puget:Sym}
Puget, J.F.:
\newblock On the satisfiability of symmetrical constrained satisfaction
  problems.
\newblock In Komorowski, J., Ras, Z., eds.: Proceedings of ISMIS'93. LNAI 689,
  Springer-Verlag (1993)  350--361

\bibitem{ssat2001}
Shlyakhter, I.:
\newblock Generating effective symmetry-breaking predicates for search
  problems.
\newblock In: Proceedings of LICS workshop on Theory and Applications of
  Satisfiability Testing (SAT 2001). (2001)

\bibitem{ffhkmpwcp2002}
Flener, P., Frisch, A., Hnich, B., Kiziltan, Z., Miguel, I., Pearson, J.,
  Walsh, T.:
\newblock Breaking row and column symmetry in matrix models.
\newblock In: 8th International Conference on Principles and Practices of
  Constraint Programming (CP-2002), Springer (2002)

\bibitem{llconstraints06}
Law, Y., Lee, J.:
\newblock {Symmetry Breaking Constraints for Value Symmetries in Constraint
  Satisfaction}.
\newblock Constraints \textbf{11}(2--3) (2006)  221--267

\bibitem{wecai2006}
Walsh, T.:
\newblock Symmetry breaking using value precedence.
\newblock In: Proc. of the 17th European Conference on Artificial Intelligence
  (ECAI-2006), European Conference on Artificial Intelligence, IOS Press (2006)

\bibitem{wcp06}
Walsh, T.:
\newblock General symmetry breaking constraints.
\newblock In: 12th International Conference on Principles and Practices of
  Constraint Programming (CP-2006), Springer-Verlag (2006)

\bibitem{wcp07}
Walsh, T.:
\newblock Breaking value symmetry.
\newblock In: 13th International Conference on Principles and Practices of
  Constraint Programming (CP-2007), Springer-Verlag (2007)

\bibitem{llwycp07}
Law, Y.C., Lee, J., Walsh, T., Yip, J.:
\newblock Breaking symmetry of interchangeable variables and values.
\newblock In: 13th International Conference on Principles and Practices of
  Constraint Programming (CP-2007), Springer-Verlag (2007)

\bibitem{waaai2008}
Walsh, T.:
\newblock Breaking value symmetry.
\newblock In Fox, D., Gomes, C., eds.: Proceedings of the 23rd National
  Conference on AI, Association for Advancement of Artificial Intelligence
  (2008)  1585--1588

\bibitem{knwcp10}
Katsirelos, G., Narodytska, N., Walsh, T.:
\newblock Static constraints for breaking row and column symmetry.
\newblock In: 16th International Conference on Principles and Practices of
  Constraint Programming (CP-2010), Springer-Verlag (2010)

\bibitem{clgrkr96}
Crawford, J., Luks, G., Ginsberg, M., Roy, A.:
\newblock Symmetry breaking predicates for search problems.
\newblock In: Proceedings of the 5th International Conference on Knowledge
  Representation and Reasoning, (KR '96). (1996)  148--159

\bibitem{snakelex}
Grayland, A., Miguel, I., Roney-Dougal, C.:
\newblock Snake lex: An alternative to double lex.
\newblock In Gent, I.P., ed.: Proceedings of 15th International Conference on
  Principles and Practice of Constraint Programming. Volume 5732 of Lecture
  Notes in Computer Science., Springer (2009)  391--399

\bibitem{bscade92}
Benhamou, B., Sais, L.:
\newblock Theoretical study of symmetries in propositional calculus and
  applications.
\newblock In: Proceedings of 11th International Conference on Automated
  Deduction. Volume 607 of Lecture Notes in Computer Science., Springer (1992)
  281--294

\bibitem{bsjar94}
Benhamou, B., Sais, L.:
\newblock Tractability through symmetries in propositional calculus.
\newblock Journal of Automated Reasoning \textbf{12}(1) (1994)  89--102

\bibitem{backofen:Sym}
Backofen, R., Will, S.:
\newblock Excluding symmetries in constraint-based search.
\newblock In Jaffar, J., ed.: Proceedings of the 5th International Conference
  on Principles and Practice of Constraint Programming. Number 1713 in Lecture
  Notes in Computer Science, Springer-Verlag (1999)  73--87

\bibitem{sbds}
Gent, I., Smith, B.:
\newblock Symmetry breaking in constraint programming.
\newblock In Horn, W., ed.: Proceedings of ECAI-2000, IOS Press (2000)
  599--603

\bibitem{dynamiclex}
Puget, J.F.:
\newblock Dynamic lex constraints.
\newblock In Benhamou, F., ed.: 12th International Conference on the Principles
  and Practice of Constraint Programming (CP 2006). Volume 4204 of Lecture
  Notes in Computer Science., Springer (2006)  453--467

\bibitem{kwecai08}
Katsirelos, G., Walsh, T.:
\newblock Dynamic symmetry breaking constraints.
\newblock In: Proceedings of the 18th ECAI, European Conference on Artificial
  Intelligence, IOS Press (2008) (under review).

\bibitem{kwecai10}
Katsirelos, G., Walsh, T.:
\newblock Symmetries of symmetry breaking constraints.
\newblock In: Proc. of the 19th European Conference on Artificial Intelligence
  (ECAI-2010), European Conference on Artificial Intelligence, IOS Press (2010)

\bibitem{cjjpsconstraints06}
Cohen, D., Jeavons, P., Jefferson, C., Petrie, K., Smith, B.:
\newblock Symmetry definitions for constraint satisfaction problems.
\newblock Constraints \textbf{11}(2--3) (2006)  115--137

\bibitem{lex2001}
Flener, P., Frisch, A., Hnich, B., Kiziltan, Z., Miguel, I., Pearson, J.,
  Walsh, T.:
\newblock Symmetry in matrix models.
\newblock Technical Report APES-30-2001, APES group (2001) Presented at
  SymCon'01 (Symmetry in Constraints), CP2001 post-conference workshop.

\bibitem{fhkmwcp2002}
Frisch, A., Hnich, B., Kiziltan, Z., Miguel, I., Walsh, T.:
\newblock Global constraints for lexicographic orderings.
\newblock In: 8th International Conference on Principles and Practices of
  Constraint Programming (CP-2002), Springer (2002)

\bibitem{fhkmwaij06}
Frisch, A., Hnich, B., Kiziltan, Z., Miguel, I., Walsh, T.:
\newblock Propagation algorithms for lexicographic ordering constraints.
\newblock Artificial Intelligence \textbf{170}(10) (2006)  803--908

\bibitem{knwercim09}
Katsirelos, G., Narodytska, N., Walsh, T.:
\newblock Combining symmetry breaking and global constraints.
\newblock In Oddi, A., Fages, F., Rossi, F., eds.: Recent Advances in
  Constraints, 13th Annual ERCIM International Workshop on Constraint Solving
  and Constraint Logic Programming (CSCLP 2008). Volume 5655 of Lecture Notes
  in Computer Science., Springer (2009)  84--98

\end{thebibliography}

\end{document}